\documentclass[acmsmall,nonacm]{acmart}
\microtypesetup{expansion=false}
\AtBeginDocument{%
  }

\usepackage{threeparttable}   
\usepackage{rotating}         
\usepackage{pifont}           
\usepackage{graphicx}         
\usepackage{array}            

\begin{document}

\title{Deep Learning Compilers: Architecture, Optimization, and Empirical Characterization}

\author{Jinghao Wen}
\email{jwen01@villanova.edu}
\author{Ruixuan Wang}
\email{rwang8@villanova.edu}
\author{Xun Jiao}
\email{xun.jiao@villanova.edu}
\affiliation{%
  \institution{Villanova University}
  \department{DEPARTMENT OF ELECTRICAL AND COMPUTER ENGINEERING}
  \city{Villanova}
  \state{Pennsylvania}
  \country{USA}
}

\begin{abstract}
As deep learning (DL) models become increasingly complex and hardware architectures grow more heterogeneous, DL compilers play a critical role in optimizing computational efficiency and enabling scalable deployment. This paper presents a systematic analysis of DL compilers, covering their architectural evolution, compilation workflows, optimization techniques, and emerging research directions. We organize DL compiler architectures into frontend, middle-end, and backend stages, emphasizing the role of the middle-end in connecting high-level model representations with hardware-specific optimizations. We further examine representative DL compilers and recent optimization techniques across different compilation stages. Beyond architectural and technical analysis, we conduct an empirical evaluation of compiler behaviors to investigate the effectiveness of graph optimization, compilation overhead, and dynamic-shape handling. Experiments using PyTorch/TorchInductor on an NVIDIA RTX A6000 demonstrate that compiler benefits depend strongly on workload characteristics, with speedups of up to 8.21× for fusion-friendly computation graphs, while other workloads exhibit limited improvements. Our results further reveal that compilation overhead can require thousands of executions to amortize, and dynamic-shape compilation reduces cumulative shape-transition costs by approximately 70\% at the expense of modest steady-state performance degradation. By combining systematic architectural analysis with quantitative experimental evidence, this work provides insights into the design, optimization, and practical deployment of modern DL compilers.
\end{abstract}




\maketitle

\section{Introduction}

From computer vision (CV)~\cite{hassaballah2019recent} and natural language processing (NLP)~\cite{chen2024multi} to recommendation systems~\cite{gao2024causal}, autonomous driving~\cite{chen2024end} and robotics~\cite{sunderhauf2018limits}, deep learning (DL) models have become foundational in a wide range of artificial intelligence (AI) applications. DL models like convolutional neural network (CNN)~\cite{lecun1998gradient}, recurrent neural network (RNN)~\cite{rumelhart1986learning}, transformer~\cite{vaswani2017attention} and graph neural networks (GNNs)~\cite{wu2020comprehensive} are growing in size and complexity in these days, and deployed on increasingly diverse hardware platforms. 

To simplify the build and deployment of complicated DL models, DL frameworks like PyTorch~\cite{paszke2019pytorch}, TensorFlow~\cite{abadi2016tensorflow}, Caffe~\cite{jia2014caffe} and Keras~\cite{moolayil2019learn} are proposed to play a crucial role in enabling efficient model development. They provide high-level abstractions for building sophisticated models, making it easier for developers to work with layers, loss functions, and optimizers. These frameworks also handle backpropagation and gradient updates automatically. However, designed to be general-purpose, these frameworks have limited ability to fully optimize code for specialized hardware like custom accelerators~\cite{wang2019various}. Furthermore, predefined operators in frameworks restrict compiler optimizations for more targeted, hardware-tailored operations~\cite{li2019hpdl}. Computational overhead associated with abstraction layers may lead to suboptimal performance compared to highly specialized compiler optimizations.

From the development of single-core processors to multi-core processors, and now with the rapid development of DL, domain-specific architectures (DSA) have risen rapidly, such as AI chips, GPU chips, and NPU chips. As hardware technology continues to advance, it has entered a new era of computing acceleration, which hardware platforms are becoming increasingly complex and multi-layered. Modern computing acceleration platform like central processing unit (CPU), graphics processing unit (GPU)~\cite{akenine2008graphics}, tensor processing unit (TPU)~\cite{jouppi2017datacenter}, neural processing unit (NPU), digital signal processor (DSP) and custom accelerator adopt multi-layer architecture, including scalar, vector, multi-core, multi-packet, multi-rack and other levels of parallel processing capabilities. These target hardware architectures are equipped with unique performance characteristics and memory hierarchies. The implementation of DL models with appointed framework on specific hardware need to manually optimize code for higher efficiency by fully using of computation resource, which is time-consuming, engineering efforts-redundant and error-prone, and cannot scale as quickly as the development of new hardware accelerators~\cite{zhang2023compiler}. Therefore, to achieve widespread application of various DL models with diverse frameworks in heterogeneous hardware architectures, the need for efficient and simplified computation has never been greater.

\begin{figure*}[!t]
    \centering
    \includegraphics[width=\textwidth]{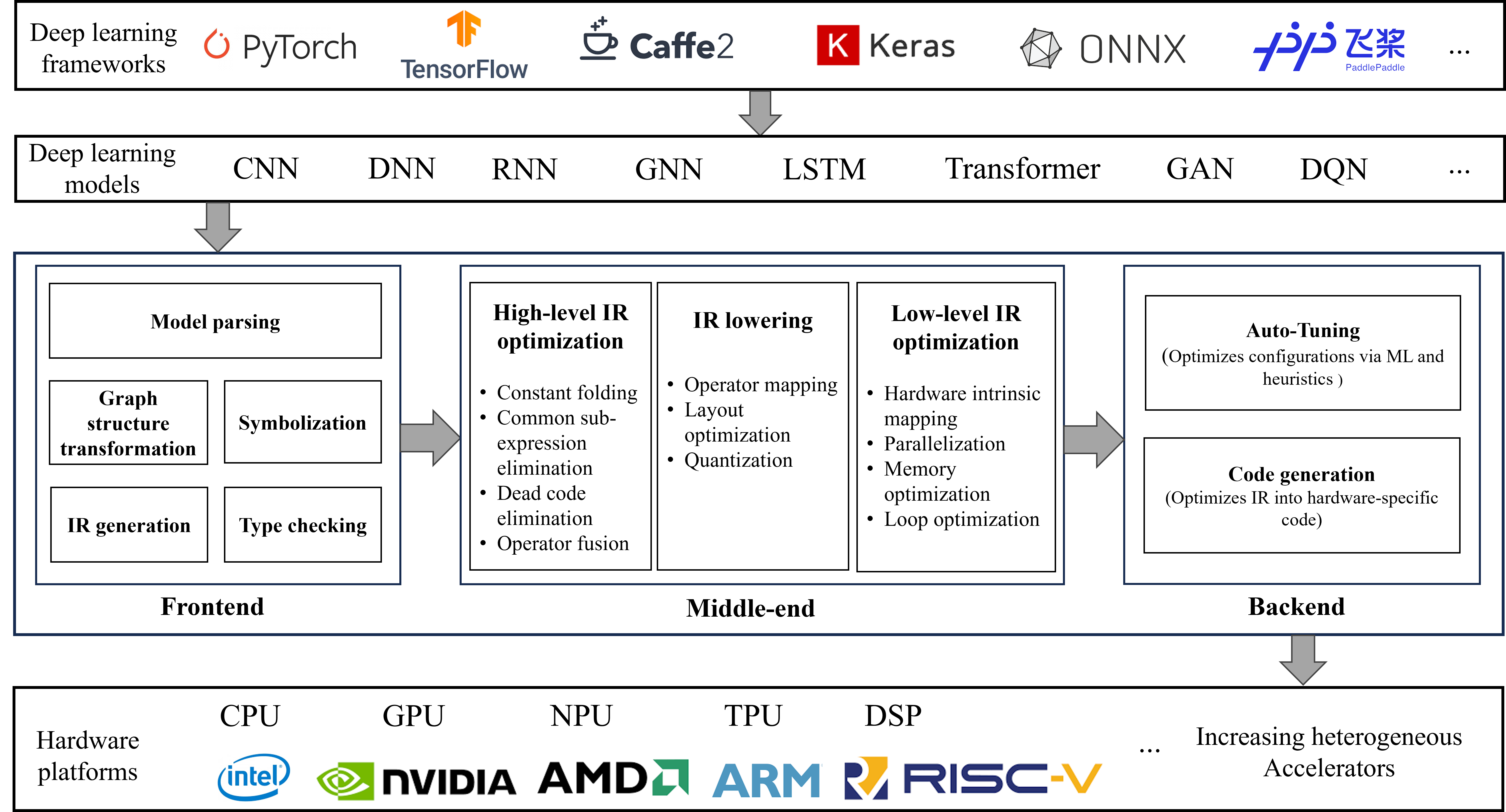}
    \caption{Normal architectural design of DL compilers.}
    \label{fig:example}
\end{figure*}

Compilers and corresponding toolsets like GNU compiler collection (GCC)~\cite{compiler2009gnu}, Clang and Low Level Virtual Machine (LLVM)~\cite{lattner2004llvm} are through the automatic way to optimize the program and code generation, so that to save a lot of manpower for different underlying hardware manual optimization~\cite{lattner2008llvm}. But these traditional compilers have limited capability to optimize DL workloads, for lacking supports of tensor operations (such as matrix multiplications, convolutions), efficient data flow management, and hardware diversity. Therefore, there is a pressing need for specialized DL compilers that can bridge the gap between high-level machine learning frameworks and efficient, hardware-specific code generation. Both industry and academia launch their own DL compilers, like Apache TVM~\cite{chen2018tvm}, Google XLA~\cite{sabne2020xla}, NVIDIA TensorRT~\cite{tensorrt}, Meta Glow~\cite{rotem2018glow}, Meta Tensor Comprehensions (TC)~\cite{vasilache2018tensor} and Intel nGraph~\cite{cyphers2018intel}, the architectural design of DL compilers is shown in Figure \ref{fig:example}. Without requiring changes to the model architecture, DL compilers improve computational efficiency, reduce memory usage, and enhance model performance by applying a series of optimizations. Key processes like transforming the model into an intermediate representation (IR), performing high and low-level optimizations, and generating hardware-specific instructions are adopted to deploy complex models at scale with improved performance, making DL compilers indispensable tools for modern ML systems~\cite{shen2021comprehensive}.

In this paper, we present a systematic analysis of deep learning compilers, focusing on their architectural evolution, compilation workflows, optimization techniques, and emerging research directions. We organize the compilation process into frontend, middle-end, and backend stages, providing a structured perspective on how high-level model representations are progressively transformed into hardware-specific implementations. In addition to examining representative DL compilers and recent optimization techniques, we investigate the practical performance implications of compiler optimizations through controlled experiments. Although modern DL compilers employ sophisticated techniques such as operator fusion, automatic scheduling, and dynamic-shape specialization, their effectiveness depends on workload characteristics, compilation overhead, and execution conditions. To better understand these factors, we evaluate representative computation graphs and analyze the trade-offs among execution latency, memory consumption, compilation cost, and shape flexibility. By combining architectural analysis with empirical validation, this work provides both a systematic understanding of DL compiler technologies and quantitative insights into their practical effectiveness.

Specifically, this paper makes the following contributions:
\begin{itemize}
\item Architectural Analysis of DL Compilers. We systematically examine the evolution of DL compilers and their architectural distinctions from traditional compilers. We organize the compilation workflow into frontend, middle-end, and backend stages, emphasizing the role of intermediate representations (IRs) and the middle-end in connecting high-level model abstractions with low-level hardware optimizations.
\item Analysis of Optimization Techniques and Recent Advances. We investigate optimization methods across different stages of DL compilation, covering representative industrial and research compilers, including TVM, XLA, TensorRT, PyTorch 2, LADDER, Welder, Cortex, and oneDNN Graph Compiler. We further discuss recent techniques such as DNNFusion, Apollo, Ansor, Relay, and SmartMem, highlighting their optimization objectives and potential applications in modern compilation workflows.
\item Empirical Validation of Compiler Behaviors. We conduct controlled experiments using PyTorch/TorchInductor to characterize compiler performance across five representative computation graphs. Our evaluation examines steady-state execution latency, memory consumption, compilation overhead, numerical agreement, and dynamic-shape handling. The results demonstrate workload-dependent speedups of up to 8.21×, substantial compilation amortization costs, and a trade-off between dynamic-shape flexibility and static specialization.
\item Emerging Challenges and Research Directions. We investigate promising directions for DL compiler development, including cross-framework interoperability, machine learning-driven optimization, debugging and verification, privacy-preserving compilation, and the integration of large language models. Together with the empirical findings, these discussions highlight important challenges and opportunities for improving the efficiency, adaptability, and reliability of future DL compiler systems.
\end{itemize}

The remainder of this paper is organized as follows. Section 2 reviews the development of DL compiler research and related studies. Section 3 introduces the evolution of traditional and DL compilers. Section 4 describes the architectural design and compilation process of DL compilers, including the frontend, middle-end, backend, and multi-level intermediate representations. Sections 5 and 6 discuss key compiler components, optimization methods, and recent technological advances. Section 7 presents an empirical evaluation of compiler behaviors, focusing on workload-dependent optimization benefits, compilation overhead, and dynamic-shape handling. Section 8 explores emerging research directions and open challenges in DL compiler development. Finally, Section 9 concludes the paper.

\section{Research Landscape and Study Positioning}

\begin{figure}[!t]
    \centering
    \includegraphics[width=\textwidth]{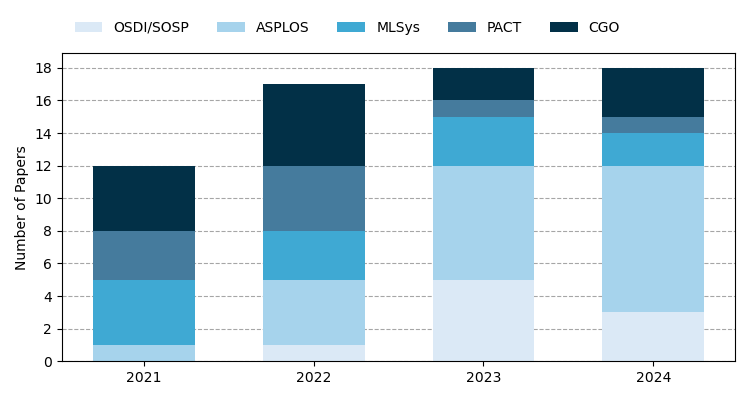}
    \caption{Quantitative summary of DL compiler publications}
    \label{fig:papers}
\end{figure}

To understand the evolution of DL compiler research, we present a quantitative summary of related publications in recent years.
Shown in Figure \ref{fig:papers}, DL compilers have garnered increasing attention from both systems and architecture research communities, as evidenced by the rising number of related publications in top-tier conferences such as OSDI/SOSP, ASPLOS, MLSys, PACT, and CGO over past years. The overall trend shows a steady increase in research activity, particularly at MLSys and ASPLOS, highlighting the field’s growing importance at the intersection of systems, compilers, and machine learning.

Table~\ref{tab:prior_work_compare} positions this work against prior DL compiler studies using their analytical coverage: compiler architecture, optimization stages, correctness, security, LLM-assisted compilation, and quantitative validation. Earlier work establishes important architectural foundations, but it generally treats the frontend and backend as the dominant organizing layers and does not connect a distinct middle-end taxonomy to controlled measurements. Our study fills this gap by coupling a stage-specific design space with an executable evaluation of fusion, amortization, memory behavior, numerical agreement, and dynamic shapes.

\begin{table}[ht]
\centering
\setlength{\tabcolsep}{2pt} 
\resizebox{\linewidth}{!}{%
\begin{threeparttable}
\caption{Positioning relative to prior DL compiler studies}
\label{tab:prior_work_compare}
\renewcommand{\arraystretch}{1.2}
\begin{tabular}{|l|c|c|c|c|c|c|c|c|c|c|c|c|c|c|}
\hline
\rotatebox{50}{\textbf{Prior Work}} &
\rotatebox{50}{\textbf{Year}} &
\rotatebox{60}{\textbf{Pages \#}} &
\rotatebox{60}{\textbf{Refs \#}} &
\rotatebox{60}{\textbf{Traditional Comp.}} &
\rotatebox{60}{\textbf{DL Comp.}} &
\rotatebox{60}{\textbf{Arch. Design}} &
\rotatebox{60}{\textbf{Frontend}} &
\rotatebox{60}{\textbf{Middle-end}} &
\rotatebox{60}{\textbf{Backend}} &
\rotatebox{60}{\textbf{Bug \& Verif.}} &
\rotatebox{60}{\textbf{Security}} &
\rotatebox{60}{\textbf{LLM}} &
\rotatebox{60}{\textbf{Quant. Analysis\textsuperscript{2}}} \\
\hline
~\cite{li2020deep} & 2020 & 20 & 75 & $\times$ & \checkmark & \checkmark & \checkmark & $\times$ & \checkmark & $\times$ & $\times$ & $\times$ & $\times$  \\
\hline
~\cite{shen2021comprehensive} & 2021 & 13 & 78 & $\times$ & \checkmark & $\approx$ & $\approx$ & $\times$ & $\approx$ & \checkmark & $\times$ & $\times$ & $\times$ \\
\hline
~\cite{zhang2023compiler} & 2023 & 21 & 153 & \checkmark & \checkmark & \checkmark & \checkmark & $\times$ & \checkmark & $\times$ & $\times$ & $\times$ & $\times$ \\
\hline
\textbf{This work} & 2025 & 26\textsuperscript{1} & 108 & \checkmark & \checkmark & \checkmark & \checkmark & \checkmark & \checkmark & \checkmark & \checkmark & \checkmark & \checkmark   \\
\hline
\end{tabular}
\begin{tablenotes}
\footnotesize
\item[1] Single-column pages.
\item[2] Quantitative analysis involving count of works, frequencies and numerical assessments.
\end{tablenotes}
\end{threeparttable}
} 
\end{table}

\section{Development of compiler}

\subsection{Background}
\noindent In the early days of computer development, programs are written with binary instructions, which are all in binary form and could be recognized directly by the CPU without any obstacles. However, they are difficult to remember and use for programmers. When the program is large, it is troublesome to write, requiring frequent reference to the instruction manual, and exceptionally frustrating to debug, as one has to directly face the confusing binary data. Additionally, binary instruction programming is a complicated process, requiring consideration of various boundary conditions and low-level issues, resulting in extremely low development efficiency. 

To improve the efficiency of program development by avoiding writing binary instructions directly, programming languages such as assembly, C, C++, Java and Python are developed to replace binary code with a simplified method. However, high-level programming languages can not be executed by hardware directly. Therefore, with the rapid development of programming languages, a special software program-compiler is proposed to transfer the programming languages to the low level binary code. Compiler can recognize the words, sentences, and various specific formats and data structures in the code of a high-level language program, and convert them into binary code, this process is called compilation~\cite{alfred2007compilers}. The invention of compiler allow developers to write code in abstract, human-readable languages, improving productivity, portability, and making software development accessible to a broader audience.

\begin{figure}[!t]
    \centering
    \includegraphics[width=\columnwidth]{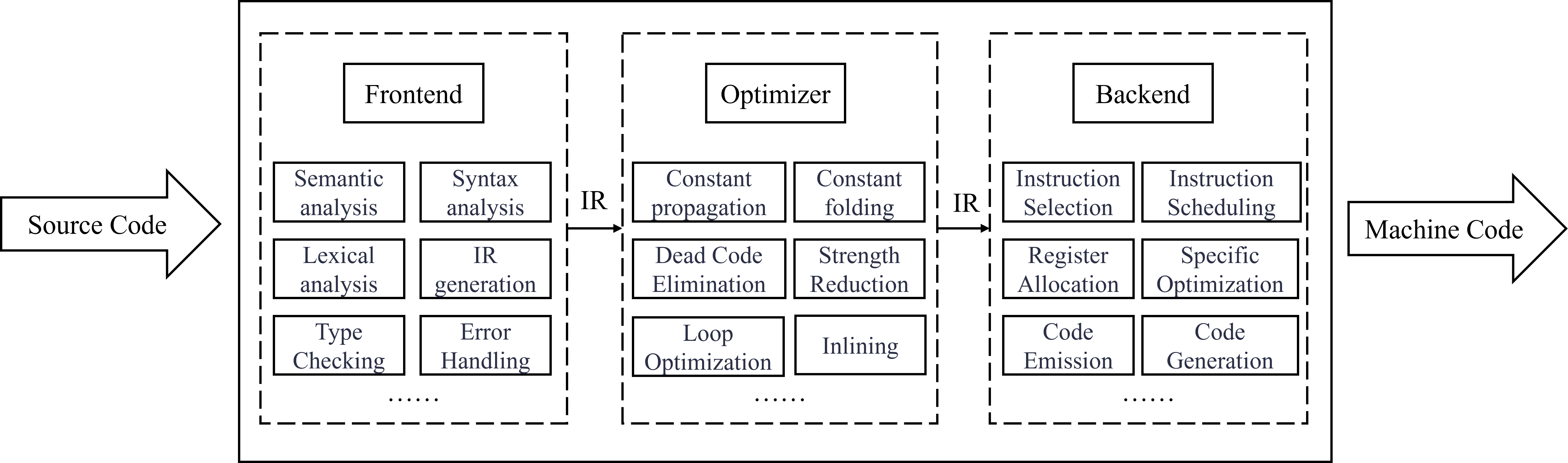}
    \caption{Normal architecture of traditional compilers.}
    \label{fig:traditional}
\end{figure}

\subsection{Traditional compilers}
\noindent Traditional compilers are essential tools in computer, translating high-level programming languages into machine code that can be executed by hardware. This process involves several stages, including lexical analysis, syntax analysis, semantic analysis, optimization, and code generation. The primary goal of a traditional compiler is to produce efficient machine code that maximizes performance while ensuring correctness~\cite{waite2012compiler}. The architecture of traditional compilers is shown in Figure \ref{fig:traditional}.

From the beginning of computer architecture to the present, after more than 60 years, compilers have developed many generations. Back in 1957, programming is done in assembly language, which causes difficult programming for most developers. Without sacrificing performance, Fortran is invented to use programs written in the high-level programming language to provide the closest performance possible to manually tuned machine code~\cite{backus1998history}. Over time, compilers evolved to include sophisticated optimization techniques, improving the execution speed of programs on various hardware architectures. These optimizations, such as loop unrolling, register allocation, and instruction scheduling, enable traditional compilers to efficiently manage memory, processor usage, and computational resources. 

Classic open source traditional compilers such as GCC, Clang and LLVM are currently mainstream for converting high-level language code into machine language code. They differ in terms of compilation speed, optimization capabilities, and supported languages, but they are all composed of three parts: the frontend, the optimizer, and the backend~\cite{lattner2008llvm}. In the internal structure of the compiler, pass is a process or module of the middle layer of compilation optimization, and the intermediate representation (IR) is responsible for connecting the layers and modules within the compiler. In traditional compiler workflows, the frontend focuses on deep semantic analysis, syntax analysis, and lexical analysis of high-level programming languages to convert source code into intermediate representation (IR). In the optimizer phase, the compiler performs a series of optimization passes that specifically target high-level language code for performance improvements. On the backend, the compiler is responsible for instruction selection, instruction scheduling, register allocation and other tasks.

\begin{figure}[t]
    \centering
    \includegraphics[width=\columnwidth]{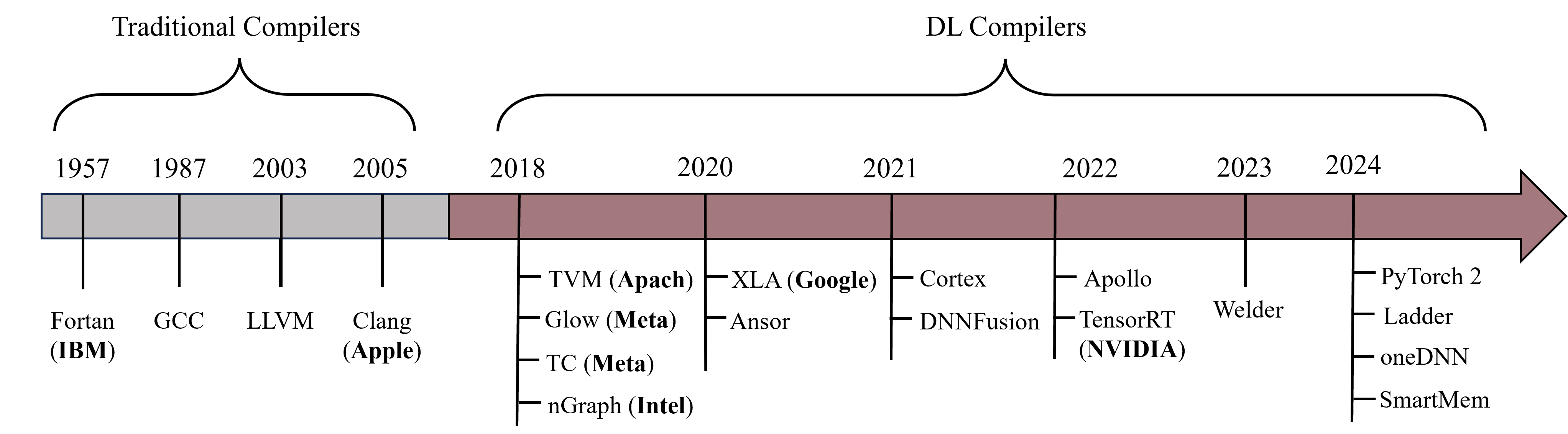}
    \caption{Timeline of compiler development.}
    \label{fig:timeline}
\end{figure}

\subsection{DL compilers}
\noindent The major limitation of traditional compilers in today's computing landscape lies in their lack of optimization for domain-specific tasks. With the rise of deep learning (DL), big data processing, and other computationally intensive workloads, traditional compilers often struggle to fully optimize for the unique requirements of these tasks. In response to the requirements, DL compiler is proposed. The timeline of compiler development is shown in Figure \ref{fig:timeline}.

The development of DL compilers has evolved through three key stages: the naive stage, the specific stage, and the general stage, as shown in Figure \ref{fig:develop}. 
The naive stage is characterized by early DL compilers, such as those in TensorFlow’s early versions, which relied on static computation graphs and operator-level abstractions for optimization ~\cite{abadi2016tensorflow}. While enabling hardware-independent and hardware-specific optimizations, such as expression simplification and operator fusion, this stage faced challenges due to rigid static graphs and limited adaptability to dynamic workloads ~\cite{yu2018dynamic}. 
The specific stage introduced AI-dedicated compilers, optimizing for deep learning workloads with enhanced flexibility and performance. DL compilers such as TVM, Tensor Comprehensions, Glow, nGraph, and XLA laid foundational work in model optimization and code generation. 
However, these compilers still faced limitations in generalization, requiring significant manual tuning for hardware-specific optimizations. 
The general stage envisions a fully automated, hardware-agnostic DL compiler capable of unified graph and operator-level optimizations, automatic code generation, and dynamic shape adaptation ~\cite{lattner2021mlir, ragan2013halide}. By integrating modular design and intelligent optimization techniques, general AI compilers aim to provide a highly flexible, efficient, and scalable compilation framework for diverse deep learning applications, spanning edge devices to data centers. Generally, the DL compilers have following typical features: 

\begin{figure}[!t]
    \centering
    \includegraphics[width=\columnwidth]{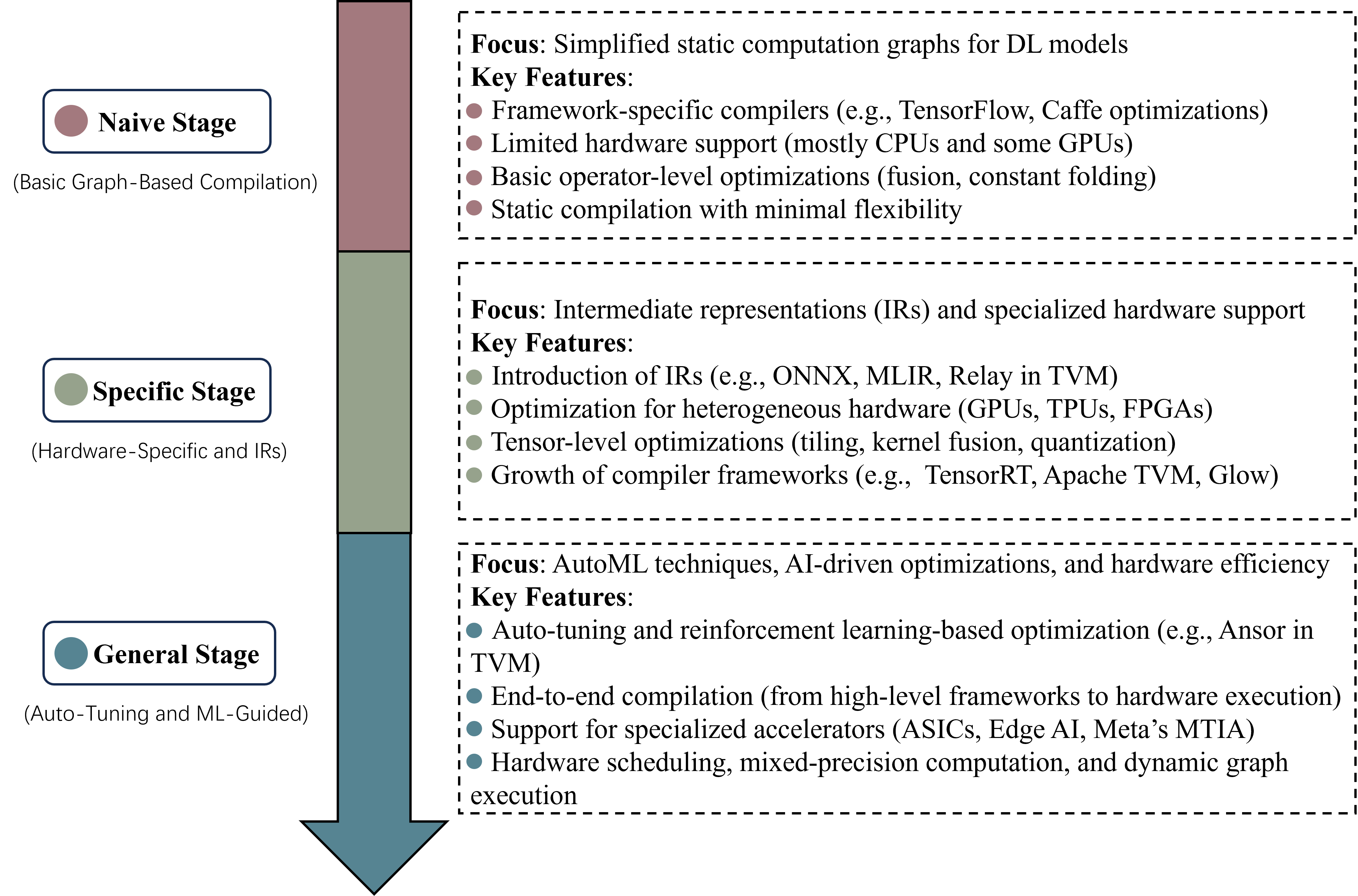}
    \caption{Three development stages of DL compilers.}
    \label{fig:develop}
\end{figure}

\begin{itemize}
\item Optimized for neural network and deep learning to achieve more efficiency in processing neural network computing tasks(e.g., automatic differentiation, gradient descent, parallel computation scheduling, model pruning and quantization).
\item Having a Python based dynamic interpreter language front-end.
\item Having multi-layer IR design, including graph compilation IR, operator compilation IR and code generation IR. 
\item Supporting for Domain-Specific Architecture (DSA) chip architecture.
\end{itemize}

\subsection{Traditional compilers VS DL compilers}
\noindent Both traditional compilers and DL compilers share the fundamental purpose of translating high-level programming language into machine code for hardware execution, easing redundant efforts on manual optimization of different underlying hardware. They are divided into frontend, optimizer and backend, and the frontend and backend are decoupled by IR to enable modular representation. After the DL compiler optimizes the Graph IR, it converts the optimized IR into the traditional compiler IR, and finally relies on the traditional compiler for machine code generation. The workflow of DL compilers relies on traditional compilers, the function of DL compilers is more like a complement to traditional compilers. 
Table \ref{table:compare} lists the comparisons between traditional compilers and DL compilers.

\begin{table}[htbp]
    \centering
    \caption{Comparison between Traditional Compilers and DL Compilers}
    \includegraphics[width=0.8\columnwidth]{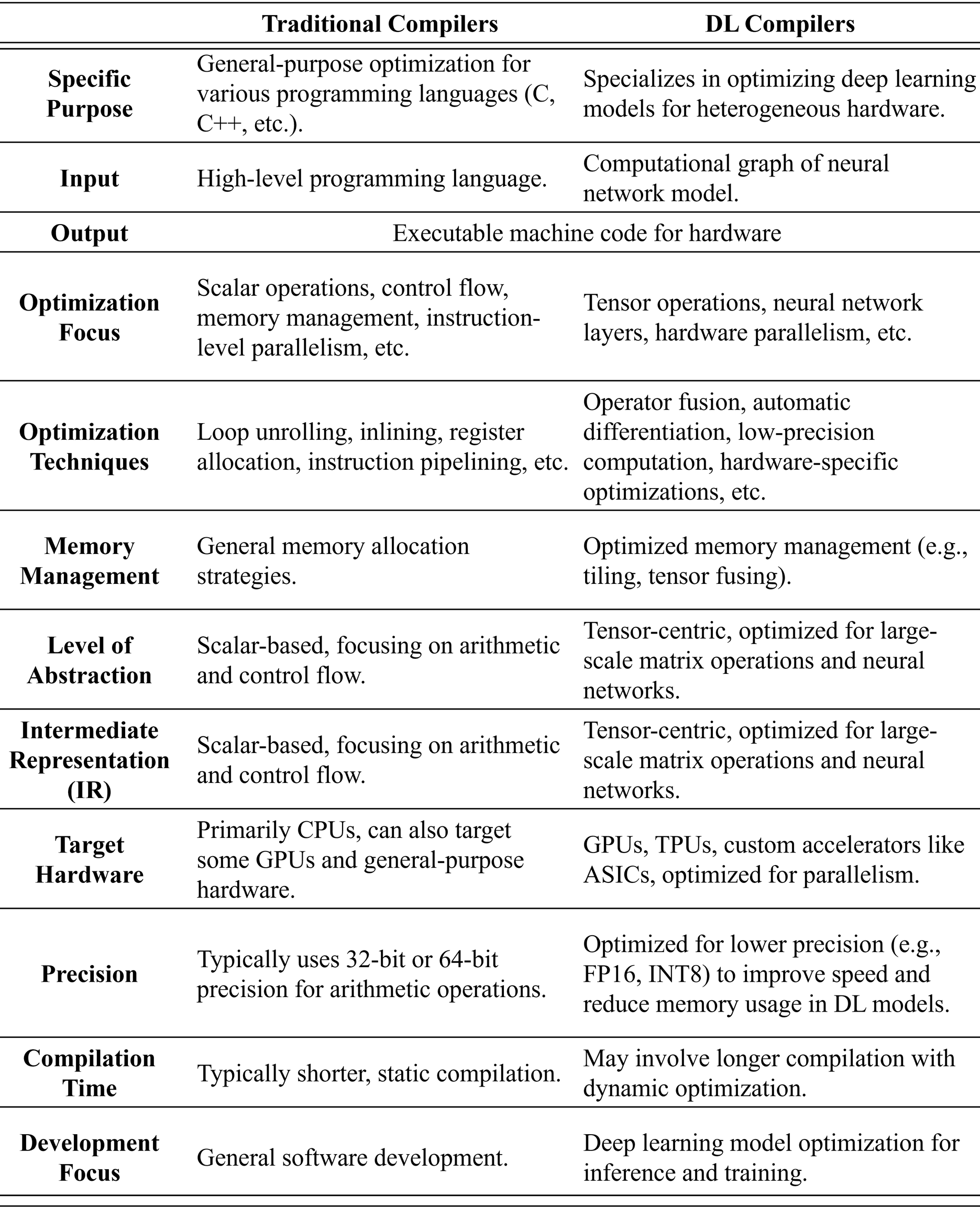}
    \label{table:compare}
\end{table}

\section{Architectural Design and Compilation Process of DL Compilers}

\subsection{Architectural Design}
\noindent The architectural design of DL compilers is tailored to efficiently handle the unique computational demands of neural networks. DL compilers typically consist of three key components: the frontend, middle-end, and backend, similar to traditional compilers but adapted for deep learning workloads. The frontend is responsible for converting high-level deep learning models, often represented in frameworks like TensorFlow or PyTorch, into a high-level IR. The middle-end performs various optimizations on both high-level IR and low-level IR, such as operator fusion, memory layout transformations, and precision tuning. Finally, the backend maps the the optimized low-level IR to specific hardware architectures, such as GPUs, TPUs, or other accelerators, exploiting parallelism and hardware-specific features to maximize performance.

\begin{table}[htbp]
    \centering
    \caption{Four abstraction compilation levels}
    \includegraphics[width=0.8\columnwidth]{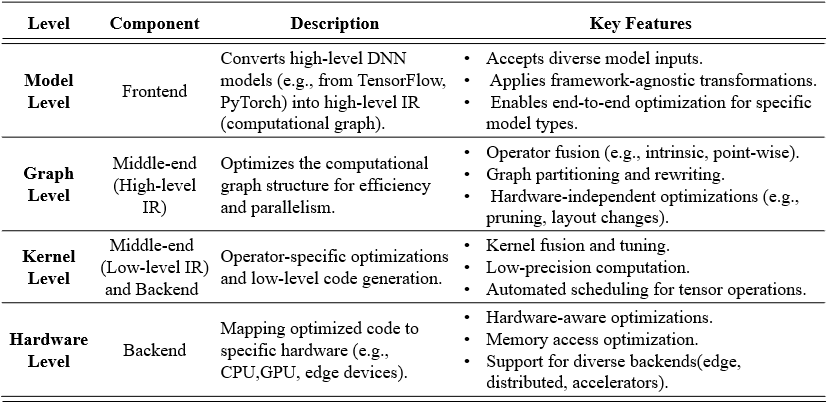}
    \label{table:level}
\end{table}

\textit{ \textbf{The frontend}} of a DL compiler is the entry point for processing high-level deep learning models. Its primary role is to translate the models, which are typically represented as computational graphs, into a high-level IR. By converting the model to IR, the frontend allows DL compilers to decouple the model logic from the underlying hardware, providing a unified interface for further optimizations. The frontend is a crucial component that bridges the gap between high-level deep learning models and the performance-optimized machine code generated by the backend.

\textit{\textbf{The high-level IR}} describes the structure of the model as a computational graph, where operations are defined as nodes and tensors as edges. It abstracts the underlying hardware and provides a framework-agnostic format that allows for a uniform optimization process later on. This level of abstraction captures operations such as matrix multiplications, convolutions, and activation functions, representing the high-level logic of the deep learning model.

\textit{\textbf{The low-level IR}} is closely aligned with the hardware's computational units and memory architecture, optimizing for parallelism, memory usage, and performance. During the optimization process in the middle-end, the high-level IR is progressively refined and transformed into a more detailed low-level IR. The transition from high-level IR to low-level IR allows the DL compiler to fine-tune the model for the specific computational capabilities of the target device. 

\textit{\textbf{The middle-end}} implements optimizations to improve the computational efficiency of the model, after the model has been transformed to high-level IR. Optimizations such as operator fusion (combining multiple operations into one), loop unrolling, memory access optimization, and precision reduction (e.g., converting FP32 to FP16 or INT8) are applied. The middle-end works on both high-level and low-level IR, progressively refining the model to better fit the hardware's characteristics. 

\textit{\textbf{The backend}} translates the optimized low-level IR into machine code that can be executed by the target hardware, which ensures that the model leverages the full capabilities of the underlying hardware, whether it's a GPU, TPU, FPGA, or ASIC. The backend generates code that takes advantage of hardware-specific features, such as parallel execution, shared memory, and cache hierarchies. Additionally, it manages the scheduling of operations and data movement between different memory hierarchies (e.g., global, shared, and register memory on GPUs). 

\subsection{Compilation Process}
\noindent Deep learning compilers process neural network workloads through a layered approach that spans multiple abstraction levels: the model level, graph level, kernel level, and hardware level. Each level corresponds to a stage in the compilation pipeline, with the frontend, middle-end, and backend interacting across these levels to transform high-level models into hardware-optimized executions, as shown in Table \ref{table:level}.

\section{Components and optimization in DL Compilers}
\noindent This section defines the major components of the DL compiler design space and associates each component with concrete optimization mechanisms. The corresponding systems and techniques are summarized in Table \ref{table:refer}.

\begin{table}[h]
    \begin{center}
    \caption{Key components and optimizations in DL compilers.}
    \label{table:refer}
    \renewcommand{\arraystretch}{1}

    \begin{tabular}{cll}
        \hline
        \textbf{Components} & \textbf{Optimization Technique} & \textbf{Reference}\\
        \hline
        \hline
        \textbf{} & 
        Model parsing&~\cite{abadi2016tensorflow, volkov2016understanding, chen2018tvm, chen2016training, nguyen2019machine, liu2020enhancing, openja2022empirical, looks2017deep} \\
        & Symbolization&~\cite{haghighat1995symbolic, wei2021llsc, lattner2021mlir, smolensky2012symbolic, kleyko2022vector} \\
        \textbf{Front-end}& Graph structure transform&~\cite{ansel2024pytorch, kim2022cprune, li2021npas, brauckmann2020compiler, boemer2019ngraph, ye2020deep, verma2022towards, guan2022dynagraph, yu2023towards, wongsuphasawat2017visualizing, hattori2020semi, gharibi2018code2graph, fegade2021cortex} \\
        & Type Checking&~\cite{shen2021comprehensive, hellendoorn2018deep, cummins2018compiler, liu2023nnsmith, shen2021nimble, verma2020shapeflow, jhoo2022static} \\
        & IR generation&~\cite{zhang2023compiler, xing2019depth, shen2021comprehensive, rotem2018glow, xing2019depth, li2020deep, roesch2019relay, majumder2023hir, lattner2021mlir, cyphers2018intel} \\
        \hline
        \textbf{} & 
        High-level IR optimization&~\cite{zheng2023operator, niu2021dnnfusion, snider2023operator, zhao2022apollo, cai2022optimus, yi2022optimizing} \\
        \textbf{Middle-end} & IR lowering&~\cite{zheng2020ansor, rotem2018glow, niu2024smartmem, liu2021post, nagel2020up, kundu2023learning, wang2024ladder} \\
        & Low-level IR optimization&~\cite{shi2023welder, li2024onednn, li2023combined, zheng2020flextensor, huang2023alcop, tavarageri2021ai, eggers1997simultaneous, volkov2016understanding} \\
        \hline
        \textbf{Back-end} & Auto-Tuning&~\cite{phothilimthana2021flexible, zheng2020ansor, zheng2020flextensor, li2020adatune, adams2019learning, ahn2020chameleon, kirkpatrick1983optimization, chen2016xgboost, phothilimthana2024tpugraphs, kaufman2021learned} \\
        & Code generation &~\cite{tillet2019triton, katel2022mlir} \\
        \hline
        \hline
    \end{tabular}
    \end{center}
\end{table}

\subsection{The frontend}
\noindent The frontend is responsible for parsing the model representation from High-Level deep learning frameworks and converting it into a high-level IR. This process contains five stages: \textit{1)} model analysis, \textit{2)} symbolization, \textit{3)} graph structure transformation, \textit{4)} type checking and optimization, \textit{5)} IR generation. The frontend is the entrance of the compiler, mainly for model translation and preliminary optimization.

\subsubsection{\textbf{Model parsing}}
 Model parsing is one of the core tasks of the frontend phase of the DL compiler, which translates the input high-level deep learning model into the computation graph. The same model can have different representation formats in different frameworks, as these frameworks have their own unique graph formats and operator definitions~\cite{nguyen2019machine, liu2020enhancing}. The model files in these different formats are read and parse to extract the core structure of the model-the computation graph~\cite{abadi2016tensorflow,volkov2016understanding}. The graph is composed of nodes (representing operations) and edges (representing data flows), and the connections between nodes represent the execution sequence and data transfer of the model~\cite{chen2018tvm, looks2017deep}. Not only the forward propagation but also the back propagation computation graph is established. The frontend automatically differentiates each parameter by tracking the data flow in the computation graph to obtain the corresponding gradient, so that these gradients can be used in the subsequent computation to update parameters in the training process of the neural network. PyTorch's "checkpoint" and JAX adopt gradient checkpointing method, storing some intermediate results temporarily in memory during forward propagation rather than keeping them all the time, to reduce the overhead and memory requirements of gradient computation ~\cite{chen2016training}. Furthermore, the frontend also recognizes the specific type of each operation and its parameter configuration, such as the convolution kernel size, the activation function type, or the size of the pooled window. Since operator definitions may differ from framework to framework, the frontend needs to map these frame-specific operators to a standard operator representation inside the compiler to achieve a uniform representation of the computation graph. 

\subsubsection{\textbf{Symbolization}} Symbolization establishes a unified interface between the compiler and the model computation graph by converting variables, tensors, and operators in the model into abstract symbolic representations ~\cite{haghighat1995symbolic}. Symbolic processing maps each input, output, and intermediate computation of the model into symbolic tensors that contain information about the type, shape, and dimension of the data, allowing the compiler to manipulate and optimize operations of the model regardless of hardware ~\cite{wei2021llsc, smolensky2012symbolic}. In addition, it maps each operation of the model (e.g., addition, convolution, or pooling, etc.) to symbolic operators that describe the computational logic, input-output dependencies, and their parameter configurations.
Through the symbolic processing, the compiler can operate the models of different frameworks on the same set of abstract representations, so that the analysis, optimization and transformation of computation graph can be carried out without relying on a specific framework.

\subsubsection{\textbf{Graph structure transformation}} Computation graphs generated by different frameworks may have different structures and definitions, and compilers need to standardize the format of computation graphs to ensure efficient processing in subsequent stages. The compiler converts the dynamic graph (e.g., PyTorch) to a static graph (e.g., TensorFlow) to unify the representation, which combines the step-by-step build operations in the dynamic graph into a single graph structure, optimizing compilation efficiency and code generation~\cite{brauckmann2020compiler, boemer2019ngraph, verma2022towards}. To optimize dynamic graphs, PyTorch 2.0 released ``TorchDynamo'', a new component that converts Python code into efficient graph mode in real time~\cite{ansel2024pytorch}. "TorchScript" is also proposed by Pytorch to transfer dynamic model to a static representation. During graph structure conversion, the compiler also breaks down complex operators (e.g., BatchNorm) into basic operators (e.g., addition, multiplication, and normalization) to streamline the graph structure and reduce dependencies. In addition, the compiler performs graph pruning, removing redundant nodes from the computed graph, to reduce the size of the graph. A graph pruning approach in deep learning compilers is used to optimize DNN execution by removing non-essential components based on target-device performance~\cite{kim2022cprune}. A compiler-aware framework is proposed to unify network pruning and architecture search to optimize DNN models, reducing computation by pruning redundant weights and operations~\cite{li2021npas}. Cortex transforms recursive control flow into an iterative form, making it suitable for static optimization, and groups recursive calls into batches before execution, reducing control flow overhead~\cite{fegade2021cortex}. After the transformation of the graph structure, the graph has a consistent structure and small scale, which not only provides a clear computation path for subsequent optimization and code generation, but also further improves the execution efficiency of the model.

\subsubsection{\textbf{Type checking}} In type checking, the compiler verifies the input and output types of each operation in the computed graph to ensure that the corresponding data type matches the requirements of the operator~\cite{shen2021comprehensive, hellendoorn2018deep, cummins2018compiler, liu2023nnsmith}. For example, some operators accept only floating-point data, while others may only support input of integer or boolean types. In addition, the compiler also performs tensor shape inference, which deduces the shape information of the output tensor by analyzing the input shape of each operator and its operation rules, ensuring that the tensor sizes of all compute nodes are compatible with each other. There are many works that are proposed to perform efficient and promising type checking. Nimble is an extended type system for static tracking of dynamic shapes and presents optimization passes for dynamic output shapes, allocation, and device placement~\cite{shen2021nimble}. ShapeFlow is a dynamic abstract interpreter designed to detect tensor shape incompatibility errors in TensorFlow programs, which constructs a custom shape computational graph to efficiently catch shape errors without requiring code annotations or modifications~\cite{verma2020shapeflow}. An automatic static analyzer, PyTea, is proposed to detect tensor shape errors in PyTorch code, it statically traces execution paths, collects tensor shape constraints, and identifies unsatisfiable constraints that could lead to shape errors during training~\cite{jhoo2022static}. Type checking not only improves the simplicity and execution efficiency of the computation graph, but also provides stable and accurate input conditions for the subsequent intermediate optimization and code generation.

\subsubsection{\textbf{IR generation}} The computation graph that has been parsed, symbolized, and structural transformation would be converted into IR, an abstract representation that can be optimized within the compiler ~\cite{zhang2023compiler,xing2019depth}. IR is designed to provide a uniform and efficient optimization interface for different hardware platforms, enabling compilers to span framework and hardware differences. As a hardware independent representation, IR expresses the logical and data-flow dependencies of each operation in a computation graph without going into specific implementation details, which enables IR to accommodate a variety of backend hardware ~\cite{lattner2021mlir, cyphers2018intel}.

In DL compilers, IR is usually divided into High-Level IR and Low-Level IR ~\cite{rotem2018glow,xing2019depth}. High-level IR is an abstract representation in the compiler close to the logic of the model itself, which preserves the high-level operators and their dependencies in the model's computation graph, often directly mapping to the operators of deep learning frameworks. Low-level IR is closer to the underlying hardware, containing specific operational implementation and data layout details. The frontend first generates high-level IR and transmits it to the middle-end, in where the high-level IR is lowered to low-level IR and delivered to the backend. This layered design makes the compiler more flexible in handling different levels of optimization tasks, facilitates cross-platform execution, and can improve performance on different hardware architectures. In the process of IR generation, the compiler will map each operation to the corresponding IR operation node according to the structure and symbolic information of the computation graph, and build a complete IR graph through the dependency relationship between the nodes, which lays a foundation for the next middle-end optimization, such as tensor fusion, operator rewriting and loop expansion optimization techniques. 

\subsection{The middle-end}
\noindent The middle-end of the DL compiler is an important phase of processing and optimizing the IR, which converts the initial high-level IR generated by the frontend into a highly optimized low-level IR suitable for the backend. The workflow at the middle-end mainly includes \textit{1)} high-level IR optimization, \textit{2)} IR lowering and \textit{3)} low-level IR optimization to improve the execution efficiency of the model and optimize the use of hardware resources.

\subsubsection{\textbf{High-level IR optimization}} By focusing on optimizations at a more abstract level, high-level IR enables compilers to perform global transformations and hardware-independent optimizations on computation graphs that improve overall model performance and scalability. Key techniques in high-level IR optimization include constant folding, common sub-expression elimination (CSE), dead code elimination (DCE) and operator fusion.

\textit{Constant folding}. To improve runtime efficiency, DL compiler evaluates constant expressions during the compilation phase, rather than during inference. Specifically, constant folding detects operations within a model that only involve constant values (like a fixed weight matrix or predefined scalars) and computes these results in advance. So that the compiler can replace these operations with precomputed values, reducing the number of operations required at runtime and thus speeding up inference.

\textit{Common sub-expression elimination}. When DL compiler detects that the same sub-expression is computed multiple times with identical inputs, CSE removes the redundancy by computing it only once and reusing the result. This reduces computation load, lowers memory usage, and can lead to faster inference times in the compiler.

\textit{Dead code elimination}. DCE is an optimization technique in DL compiler that removes parts of a model’s computation graph that do not contribute to the final output. This optimization focuses on finding and eliminating "dead" or unused computations, which include operations, variables, and nodes that are not required for the model’s inference or backpropagation.

\textit{Operator fusion}. By combining multiple operators into a single operation, DL compilers can reduce overhead related to intermediate memory storage and streamline computation ~\cite{yi2022optimizing}. TVM recognizes four categories of graph operators: injective, reduction, complex-out-fusible, and opaque, and proposes generic rules to fuse operators. A scheduling rule for fusing memory-intensive operators is furthermore designed, by assigning scheduling policies to arithmetic using scheduling primitives ~\cite{zheng2023operator}. A novel and extensive loop fusion framework, DNNFusion, optimizes operator execution by combining graph rewriting and mathematical transformation techniques ~\cite{niu2021dnnfusion}. XLA employs several fusion strategies, such as instruction fusion, horizontal fusion, and multi-output fusion, showing significant improvements in both TensorFlow and JAX workflows~\cite{snider2023operator}. TensorRT aggressively fuses operators to minimize memory usage and computation overhead, optimizing for NVIDIA GPUs~\cite{tensorrt}. Apollo, designed specifically for scalable fusion in DNN compilers, incorporates a multi-layer fusion mechanism. It uses polyhedral loop fusion for dependence-based fusion and explores parallelism stitching across operations ~\cite{zhao2022apollo}. Optimus generates high-efficiency operator-fusion solutions to guarantee the optimality of memory level in deep neural networks ~\cite{cai2022optimus}. These techniques reflect ongoing efforts to refine operator fusion to improve performance, particularly for DL models on resource-constrained devices and across various hardware architectures.

\subsubsection{\textbf{IR lowering}} The conversion of high-level IR to low-level IR is a key step in the middle-end of the compiler. At this stage, high-level abstract representations are gradually refined into concrete low-level operators, and attention is paid to the requirements and limitations of the underlying hardware implementation. This process usually involves several steps, this paper discuss three most common steps: operator mapping, layout optimization and quantization.

\textit{Operator mapping}. High-level operators, like matrix multiplication or convolution, are mapped to low-level, platform-specific instructions in low-level IR that are optimized for the target hardware. This mapping process, also known as "operator lowering," ensures that the computation graph of a deep learning model is not only compatible with the hardware but also takes advantage of the hardware’s specialized instructions and features for optimal performance.

\textit{Layout optimization}. This involves adjusting data storage formats to maximize hardware efficiency. Proper memory layout can improve cache utilization and reduce memory access latency, which is crucial for accelerating DL models. In Relay, a high-level intermediate representation in TVM, data formats are adjusted, typically using NCHW (batch-channel-height-width) or NHWC (batch-height-width-channel). Ansor, built on top of TVM, explores layout optimization by using a search space to automatically select the best layouts~\cite{zheng2020ansor}. Glow uses layout transformations to match the optimal data format for each backend, aiming to reduce memory usage and improve computational efficiency~\cite{rotem2018glow}. SmartMem enhances DNN execution on mobile devices by eliminating redundant layout transformations and dynamically adapting data layouts~\cite{niu2024smartmem}. While the actual transformations are most efficiently handled in the middle-end, frontend (like high-level APIs in PyTorch or TensorFlow) can specify layout preferences and backend (like XLA) also implements target hardware constraints on data layout.  

\textit{Quantization}. Quantization reduces the precision of model parameters from floating-point numbers to lower-precision formats (such as FP16 or INT8), making models more efficient by reducing memory usage, improving latency, and enabling faster inference. TVM generates fixed-point data types operators to make ultra low-precision inference. Profile-guided quantization in Glow estimates numeric range according to inferring execution. TensorRT supports post-training quantization to FP16 and INT8. A post-training quantization is adapted for vision transformers by applying advanced calibration methods that minimize quantization error in attention mechanisms~\cite{liu2021post}. AdaRound improves post-training quantization by learning how to adjust the rounding of weights to minimize accuracy loss~\cite{nagel2020up}. Introduced by Intel, SENet reduces the number of ReLU activations per layer based on their sensitivity, optimizing for latency and resource efficiency during inference~\cite{kundu2023learning}. A new compiler, LADDER, optimizes low-precision deep learning computation by enabling efficient tensor transformations and scheduling for custom low-bit data types~\cite{wang2024ladder}.

\subsubsection{\textbf{Low-level IR optimization}} The low-level IR optimization aims to adapt the computation graph to a specific hardware architecture, minimizing resource consumption and execution time. Low-level IR is a representation close to hardware implementation, focusing on specific optimizations such as memory, instruction parallelism, and register utilization. Optimization at this stage typically includes: hardware intrinsic mapping, parallelization, memory optimization, and loop optimization.

\textit{Hardware intrinsic mapping}. Hardware intrinsic mapping in DL compiler refers to the process of tailoring low-level IR operations to utilize specific hardware instructions or capabilities available on target devices, such as CPUs, GPUs, TPUs, or custom accelerators. This mapping optimizes performance by leveraging hardware-specific features like vectorized instructions, tensor cores, or specialized ALUs.

\textit{Parallelization}. In DL compiler optimization, parallelization indicates that the compiler utilizes the parallelism in the target hardware device to maximize the performance. Welder~\cite{shi2023welder} compiler abstracts the data management to tile-graph and optimizes the tile graph by selecting hardware-efficient tile configurations and aligning computation parallelism with GPU architecture. Meanwhile, targeting recursive DNN models, Cortex~\cite{fegade2021cortex} employs the property of recursive models that control flows in the model are encapsulated in the input data structure. Cortex utilizes dynamic batching to exploit parallelism in a batch with dynamic control flow, effectively using the higher parallelism on the GPU device.

\textit{Memory optimization}. DNN models often offload large amounts of data to on-chip memory, leading to delays due to data allocation and fetching, which can become a bottleneck in execution~\cite{gholami2024ai}. Efficient memory allocation is crucial for managing how memory is assigned, pooled, and structured to minimize fragmentation and ensure large models meet their memory requirements effectively. Memory optimization approaches typically include memory pooling, memory planning, memory reusing, and memory latency hiding.

\begin{itemize}
\item Memory pooling: Memory pooling reduces the overhead associated with frequent allocation and deallocation by setting up a pool of pre-allocated memory blocks. When tensors are no longer needed, their memory is returned to the pool and quickly reassigned to new tensors. 

\item Memory planning: Automatic memory planning algorithms are used to determine an optimal layout of tensor allocations based on the target hardware’s memory constraints.
COSMA alsominimizes off-chip memory access by co-optimizing operator scheduling and memory allocation for DL models on hardware accelerators~\cite{li2023combined}. FlexTensor performs memory-aware operator scheduling, which uses a hybrid static-dynamic memory allocation method tailored for FPGA accelerators~\cite{zheng2020flextensor}.

\item Memory reusing: Memory reuse in DL compiler minimizes peak memory requirements by reassigning memory blocks no longer in use to subsequent operations. DL compilers like TensorFlow XLA and TVM use lifetime analysis to track the scope of each tensor and free memory when a tensor data is no longer required. Relay and ONNX implement buffer sharing to avoid redundant allocations by reusing buffers among tensors that are sequentially dependent but do not overlap. For some operations that do not need to preserve the original data, compilers such as Glow use in-place computation, which allows transformations to overwrite the original tensor data. On the other hand, Welder~\cite{shi2023welder} introduces a tile-graph abstraction for fine-grained memory management, enabling unified optimization across memory reuse and placement for both accelerators and host systems~\cite{shi2023welder}. OneDNN graph compiler implements hybrid memory reuse strategies that combine dynamic and static memory allocation with buffer sharing~\cite{li2024onednn}.

\item Memory latency hiding: Moreover, to mitigate the memory access delay, memory latency hiding is also utilized in the middle end optimization. To achieve this goal, memory latency hiding overlaps memory operations with computational operations. Different hardware devices requires variant approaches for memory latency hiding. CPUs and GPUs can naturally support memory latency hiding by SMT ~\cite{eggers1997simultaneous} and rapid context switching ~\cite{volkov2016understanding}. On the other hand, specific DL accelerators can use a decoupled access-execute architecture to reduce runtime latency. 
\end{itemize}

\textit{Loop optimization}. Loop optimization involves improving the execution of loops in deep learning workloads to maximize hardware utilization and minimize computational and memory overhead. Loops are prevalent in DNN operations like matrix multiplication, convolution, and backpropagation, making their optimization critical for performance. FlexTensor dynamically schedules loops by optimizing parallel thread-level execution, achieving high computational throughput by adapting loop tiling and fusion~\cite{zheng2020flextensor}. ALCOP introduces automatic load-compute pipelining, optimizing loop nests by synchronizing memory barriers and leveraging asynchronous memory operations~\cite{huang2023alcop}. Another approach adopts polyhedral compilation and reinforcement learning to optimize loop transformations for both sequential and parallel deep learning workloads, achieving significant speed-ups by leveraging multi-level cache hierarchies and SIMD units~\cite{tavarageri2021ai}. Common loop optimization adapted by DL compiler includes loop unrolling, loop fusion, loop tiling, loop reordering and loop parallelization.

\begin{itemize}
\item Loop unrolling: Loop unrolling expands loops by executing multiple iterations in a single loop cycle, reducing loop overhead and enabling better utilization of vectorized hardware units. Compilers like LLVM or TVM unroll loops to exploit SIMD instructions on CPUs or GPUs.
\item Loop fusion: By combining multiple loops into a single loop, DL compilers like TVM, XLA and TC reduce memory accesses and improve cache locality.
\item Loop tiling: Loop tiling breaks loops into smaller "tiles" to optimize memory locality by ensuring chunks of data fit into fast memory like L1/L2 caches. XLA and FlexTensor employ tiling for convolution operations to minimize cache misses.
\item Loop reordering: Compiler changes the order of nested loops to improve memory access patterns for reducing latency. TVM performs loop reordering during scheduling to prioritize computations with better data locality.
\item Loop parallelization: Loop parallelization splits loop iterations across multiple processing units (threads or cores) to enhance parallel execution. CUDA-based parallelism is automatically inserted by compilers like Glow for supported backends.
\end{itemize}

\subsection{The backend}
\noindent Typically, the compiler backend lowers the high-level IR with different optimizations and transform the optimized IR into code for the target devices based on specific hardware configurations. For this perspective, the backend optimization methodologies can be categorized into auto-tuning, which utilizes ML model to automatically determine the optimal configuration of parameters, and code generation, generating the execution code for hardware platforms. The following of this section will dive deep into these two aspects of backend optimizations.

\subsubsection{\textbf{Auto-Tuning}}
Auto-tuning leverages automated approaches to identify the optimal configurations for hardware-specific optimization and code generation. Given the vast search space involved in configuration tuning, automatic tuning is essential to reduce manual efforts and improve search efficiency~\cite{phothilimthana2021flexible}. Most modern deep learning compilers integrate auto-tuner in the compiler backend to facilitate this process~\cite{zheng2020ansor, zheng2020flextensor, li2020adatune, adams2019learning, ahn2020chameleon}.

Generally speaking, auto-tuners consist of a ML based estimator for performance estimation and a searching algorithm capable of exploring more complex configurations. For instance, TVM implements the automated schedule optimizer considering both schedule optimizations and schedule-specific parameters optimizations. To efficiently navigate the large search space and identify the optimal configurations, TVM utilizes a heuristic algorithm named parallel simulated annealing algorithm~\cite{kirkpatrick1983optimization} combined with an XGBoost-based model~\cite{chen2016xgboost} for configuration optimization. XLA uses graph neural networks (GNNs) to predict optimal tile sizes, selecting top-K configurations for real-device execution.

Meanwhile, tree-based ML models have been widely employed in compiler backend for performance prediction. Similarly to TVM, Ansor adopts a sampling-based approach for schedule exploration and employs a gradient boosting decision tree (GBDT) as its ML-based predictor. AdaTune~\cite{li2020adatune} implements a Random Forest regressor within the backend pipeline. To further improve the optimization efficiency, AdaTune introduces an adaptive exploration process that can dynamically balance exploration and exploitation through early termination. Additionally, AdaTune dynamic adjust the searching process based on posterior distribution of configurations, which can increase the adaptivity of searcher and reducing the likelihood of suboptimal results.

Moreover, FlexTensor~\cite{zheng2020flextensor} explores optimization design spaces consisting of various schedules and utilizes reinforcement learning (RL) to identify the optimal schedule configuration. XLA takes a different approach by leveraging a graph neural networks (GNNs) model to predict the performance and otpimal configurations directly from the topological informational of high-level IRs~\cite{phothilimthana2024tpugraphs}. The auto-tuner in XLA backend mainly focus on the tile size selection, which optimize the shape of a tile of tensors to maximize the computational efficiency of hardware. To improve the performance of the tile sizing prediction, XLA employ supervised learning based on tensor operator graph and select the top-K most promising configurations to execute on real devices ~\cite{kaufman2021learned}.

On the other hand, incorporating an ML-based estimator can increase overall compilation time. To enhance the compilation time efficiency, recent proposed DL compilers have adopted hierarchical search strategies guided by hardware awareness metrics, eliminating the need for ML models in auto-tuning. For instance, Welder~\cite{shi2023welder} optimizes memory efficiency across different memory layers by using memory footprint and total memory traffic as a cost function to guide scheduling optimization. Similarly, the oneDNN graph compiler~\cite{li2024onednn} employs a two-level IR system, graph IR and tensor IR, to simplify the optimization pipeline and improve efficiency using expert-derived heuristics. Instead of relying on a costly tuning process, the oneDNN graph compiler leverages the expert-tuned templates and lower-level IRs to determine the optimal configuration.

\subsubsection{\textbf{Code generation}}
Code generation is a crucial functionality of the compiler backend. Typically the compiler backend transforms the scheduled operators from the lower-level IR into a series of function calls during the code generation process. For instance, XLA utilizes tensor-level IR transformation to map tensor program onto pre-defined LLVM-IR and CUDA operation templates based on pattern matching. In the meantime, TVM applies schedule primitives to incorporate with the schedule optimizations in tensor expressions, and lowers the scheduled operations into hardware-specific code such as LLVM IR for CPUs and CUDA kernels for GPUs.

However, in practice, compiler-generated code often underperforms compared to vendor-implemented kernels such as cuDNN and cuBLAS, though these vendor libraries support only a limited set of operations. To address this limitation, Triton~\cite{tillet2019triton} introduces Triton-JIT for efficient GPU code generation based on tile-level semantics analysis and optimization.

In addition to LLVM IR, which is widely utilized in compiler backend code generation due to its flexibility and robustness, Multi-Level IR (MLIR) based code generation ~\cite{katel2022mlir} has emerged to reduce reliance on manually tuned libraries and eliminate redundant hardware-aware optimizations. MLIR provides a transformation and lowering pipeline that automates high-performance code generation.

Moreover, Cortex introduces the Irregular Loops IR (ILIR) for an end-to-end compilation and code generation. ILIR is a loop-based, data-structure-agnostic IR that extends tensor compiler IRs to support irregular and recursive computations.  Specifically, ILIR employs bound inference and loop iteration space indexing to enhance the computational and memory efficiency. By integrating scheduling primitives, Cortex achieves highly efficient code generation through an end-to-end optimization approach.

Besides, the oneDNN Graph compiler employs a hybrid approach that combines compiler optimizations with expert-tuned kernels for high-performance code generation. In practice, it first generates operator code using expert-tuned templates and then optimizes internal parameters through heuristic algorithm.

Additionally, to address the inefficiencies and limitations of traditional compilers in mixed-precision computation field, LADDER introduces tTile, a novel IR that supports a wide range of data types with arbitrary bit widths by decoupling data storage from computation. Furthermore, as an extension of TVM code generation, the LADDER compiler directly generates lower-level parallel-thread execution (PTX) code, offering greater flexibility and improved performance compared to conventional CUDA C++ code.


\begin{table}[ht]
    \centering
    \caption{Summary of industrial DL compilers}
    \label{table:industrial_compilers}
    \renewcommand{\arraystretch}{1}
    \small
    \setlength{\tabcolsep}{3pt} 
    \resizebox{\linewidth}{!}{
    \begin{tabular}{m{1.8cm} m{1.8cm} m{3.5cm} m{3.8cm} m{3.4cm}}
        \hline
        \textbf{Name} & \textbf{Category} & \textbf{Key Features} & \textbf{Advancements} & \textbf{Target Case}\\ 
        \hline
        \hline
        \textbf{TVM}~\cite{chen2018tvm} & 
        DL compiler &  
        Relay IR, hardware abstraction, AutoTVM \& Ansor schedulers. &
        Modular backend support, ML-based autotuning, deep tensor-level optimization. & 
        Multi-platform deployment (CPU, GPU, accelerators)\\  
        \hline
        \textbf{XLA}~\cite{sabne2020xla} & 
        DL compiler &  
        HLO IR, graph-level optimizations, AOT compilation. &
        Operator fusion, memory reuse, tile-size tuning, integration in TF \& JAX. & 
        High-performance workloads on TPUs/GPUs/CPUs\\  
        \hline
        \textbf{Tensor-RT}~\cite{tensorrt} & 
        DL compiler &  
        Inference-only, FP16 and INT8 optimization, kernel fusion. &
        Layer fusion, kernel tactic selection, high-throughput inference. & 
        NVIDIA GPU inference pipelines\\  
        \hline
        \textbf{Glow}~\cite{rotem2018glow} & 
        DL compiler &  
        Two-level IR (GraphIR \& IRFunction), quantization support. &
        Memory optimization, backend pluggability, support for embedded devices. & 
        Edge inference, resource-constrained environments\\  
        \hline
        \textbf{nGraph}~\cite{cyphers2018intel} & 
        DL compiler &  
        IR modularity, multi-framework frontend, low-precision support. &
        Codegen for Intel HW, memory layout transformation, quantization-aware IR. & 
        Intel platforms (CPU, GPU, VPU)\\  
        \hline
        \textbf{TC}~\cite{vasilache2018tensor} & 
        Optimization Framework &  
        Tensor DSL, polyhedral scheduler, autotuning. &
        Automatic CUDA/OpenCL kernel generation, differentiable compilation. & 
        Custom kernel generation for DL acceleration\\  
        \hline
        \hline
    \end{tabular}
    } 
\end{table}

\section{The taxonomy of new compilers and optimizations}

\noindent We instantiate the taxonomy using industrial DL compilers such as TVM, XLA, TensorRT, Glow, TC, and nGraph, as shown in Table \ref{table:industrial_compilers}, and newer systems including LADDER, Welder, PyTorch 2, Cortex, and the oneDNN Graph Compiler. These systems occupy different points in the design space and expose distinct trade-offs in code generation, performance portability, graph capture, scheduling, and hardware specialization. For example, LADDER targets low-precision performance portability, Welder coordinates tile-level memory movement, Cortex handles recursive model structure, PyTorch 2 combines dynamic Python bytecode capture with graph compilation, and oneDNN Graph Compiler combines compiler transformations with expert-tuned kernels.

Additionally, we explore novel optimization techniques that have emerged alongside these compilers, including DNNFusion, Apollo, Ansor, Relay and SmartMem, which collectively push the boundaries of DNN execution efficiency. DNNFusion introduces an advanced operator fusion framework that expands fusion opportunities beyond traditional restrictive patterns, leveraging mathematical-property-based graph rewriting and lightweight profiling to accelerate inference by up to 9.3x compared to existing frameworks. Apollo enhances this further with its layer-by-layer, partition-based fusion strategy, optimizing computational graphs for specific hardware targets. Ansor, on the other hand, brings a sophisticated approach to tensor program generation, employing automated scheduling and exploration to produce high-performance code tailored to diverse architectures. Relay enhances flexibility with its Relay-based IR, supporting dynamic tensor shapes and optimization passes like fusion and constant folding. SmartMem introduces memory optimization strategies that reduce overhead and improve resource utilization, addressing the memory-intensive nature of modern DNNs. The summary of these new compilers and optimizations is shown in Table \ref{table:taxonomy}.


\begin{table}[ht]
    \centering
    \caption{Summary of new compilers and optimizations}
    \label{table:taxonomy}
    \renewcommand{\arraystretch}{1}
    \small
    \setlength{\tabcolsep}{3pt} 
    \resizebox{\linewidth}{!}{%
    \begin{tabular}{m{2.2cm}m{2cm}m{3.5cm}m{4.3cm}m{3cm}}  
        \hline
        \textbf{Name} & \textbf{Category} & \textbf{Key Features} & \textbf{Advancements} & \textbf{Target Case}\\ 
        \hline
        \hline
        \textbf{Ladder}~\cite{wang2024ladder} & DL compiler &  
        Hardware-aware tensor transformation for low-precision computing.&
        Optimizes low-precision data types, improves efficiency on resource-constrained devices. & 
        Edge devices, low-power hardware\\  
        \hline
        \textbf{Welder}~\cite{shi2023welder} & DL compiler &  
        Tile-graph-based memory access scheduling for DNNs.&
        Optimizes memory access patterns via tile graphs, reducing latency and improving throughput. & 
        Large-scale DNNs, memory-bound tasks\\  
        \hline
        \textbf{Pytorch 2}~\cite{ansel2024pytorch} & DL compiler &  
        Dynamic Python bytecode transformation and graph compilation.&
        Captures graphs with TorchDynamo, optimizes via TorchInductor, achieves up to 40\% training speedup. & 
        General DL workloads, PyTorch users\\  
        \hline
        \textbf{Cortex}~\cite{fegade2021cortex} & DL compiler &  
        Compiler for recursive deep learning models. &
        End-to-end optimizations for recursive models, achieves up to 14x lower inference latency. & 
        Recursive DNNs, low latency inference\\  
        \hline
        \textbf{oneDNN}~\cite{li2024onednn} & DL compiler &  
        Hybrid approach combining compiler optimizations and expert-tuned kernels. &
        Aggressive operator fusion, low-precision computation, static tensor shape optimization. & 
        High performance DNNs\\  
        \hline
        \textbf{DNNFusion}~\cite{niu2021dnnfusion} & Optimization Technique &  
        Advanced operator fusion for DNN execution. &
        Mathematical-property-based graph rewriting, integrated fusion planning, achieves up to 9.3x speedup. & 
        Mobile devices, deep DNNs\\
        \hline
        \textbf{Apollo}~\cite{zhao2022apollo}& Optimization Technique &  
        Automatic partition-based operator fusion through layer-by-layer optimization. &
        Layer-wise partitioning and fusion, enhances parallelism and reduces computation overhead. & 
        DNN models, distributed systems\\  
        \hline
        \textbf{Ansor}~\cite{zheng2020ansor}& Optimization Technique &  
        Auto-scheduling framework for high-performance tensor programs. &
        Automated schedule exploration, achieves high performance across diverse hardware. & 
        Heterogeneous systems, tensor programs\\
        \hline
        \textbf{Relay}~\cite{roesch2019relay}& Optimization Technique &  
        Relay-based IR with static and dynamic shape support and operator optimization.&
        Extends Relay IR for dynamic tensor shapes, integrates optimization passes like fusion and constant folding. & 
        Dynamic DNNs, flexible workloads\\  
        \hline
        \textbf{SmartMem}~\cite{niu2024smartmem}& Optimization Technique &  
        Intelligent memory optimization via operator-level memory replay. &
        Reduces memory footprint by up to 2.1x, leverages operator-level reuse and recomputation trade-offs. & 
        Memory-constrained DNN execution\\
        \hline
        \hline
    \end{tabular}
    }
\end{table}

Different compilers and optimizations are designed for specific scenarios, as  summarized in Table \ref{table:use}. 
For compute-intensive models requiring aggressive operator fusion, TVM, TC, oneDNN, PyTorch 2, DNNFusion, and Apollo are ideal, as they enhance execution efficiency through fusion techniques. When handling inference-only workloads that demand high performance and low latency, TensorRT, XLA, Glow, Welder, oneDNN, Cortex, PyTorch 2, DNNFusion, Apollo, and Ansor optimize inference by minimizing code changes and improving kernel execution. When working with low-precision data types, TVM, TensorRT, XLA, Ladder and oneDNN support efficient low-bit computations for better accelerator performance.  Automated optimization and scheduling are best handled by TVM, TC, Ansor and Relay, which generate high-performance execution plans. In resource-constrained environments like embedded systems and microcontrollers, TVM, Glow, Ladder, oneDNN, and SmartMem improve efficiency by optimizing low-precision computations or reducing memory usage. For distributed or multi-device systems,  XLA and Apollo improves execution efficiency through partition-based fusion. Dynamic shape models benefit from nGraph, PyTorch 2 and Relay, which enable faster compilation and flexible execution across hardware. Finally, XLA, nGraph and PyTorch 2 can support and optimize training workloads.

\begin{table}[ht]
    \centering
    \caption{Compilers and optimizations designed for specific scenarios.}
    \renewcommand{\arraystretch}{1}
    \begin{tabular}{m{7.5cm} m{5.5cm}}
        \hline
        \textbf{Scenarios} & \textbf{Compilers and optimizations} \\
        \hline\hline
        Compute-intensive models requiring aggressive operator fusion: & 
        \cite{chen2018tvm}, \cite{vasilache2018tensor}, \cite{li2024onednn}, \cite{ansel2024pytorch}, \cite{niu2021dnnfusion}, \cite{zhao2022apollo} \\
        \hline
        Inference workloads needing high performance and low latency: &
        \cite{sabne2020xla}, \cite{tensorrt}, \cite{rotem2018glow}, \cite{shi2023welder}, \cite{fegade2021cortex}, \cite{li2024onednn}, \cite{ansel2024pytorch}, \cite{niu2021dnnfusion}, \cite{zhao2022apollo}, \cite{zheng2020ansor} \\
        \hline
        Model optimizations with low-precision data types: & 
        \cite{chen2018tvm}, \cite{sabne2020xla}, \cite{tensorrt}, \cite{wang2024ladder}, \cite{li2024onednn} \\
        \hline
        Automated optimization and scheduling: &
        \cite{chen2018tvm}, \cite{vasilache2018tensor}, \cite{zheng2020ansor}, \cite{roesch2019relay} \\
        \hline
        Resource-constrained environments: & 
        \cite{chen2018tvm}, \cite{rotem2018glow}, \cite{wang2024ladder}, \cite{li2024onednn}, \cite{niu2024smartmem} \\
        \hline
        Distributed or multi-device systems: & 
        \cite{sabne2020xla}, \cite{zhao2022apollo} \\
        \hline
        For dynamic shape models: & 
        \cite{cyphers2018intel}, \cite{ansel2024pytorch}, \cite{roesch2019relay} \\
        \hline
        For training workloads: & 
        \cite{sabne2020xla}, \cite{cyphers2018intel}, \cite{ansel2024pytorch} \\
        \hline
        \hline
    \end{tabular}
    \label{table:use}
\end{table}

Together, these advancements signify a transformative shift in DL compilation, offering robust solutions to the challenges of scalability, latency, and hardware heterogeneity in the rapidly evolving landscape of AI. 
The synergy between advanced fusion methods, automated scheduling, and memory-aware optimizations allows for more efficient execution of deep learning workloads, minimizing computational bottlenecks and maximizing hardware utilization. Moreover, these innovations contribute to a broader movement toward unified compilation frameworks that seamlessly support heterogeneous accelerators. This shift is crucial as deep learning applications continue to grow in complexity, demanding greater computational efficiency without sacrificing model performance. The continuous evolution of compiler techniques underscores the field's commitment to optimizing both inference and training pipelines. As deep learning models push the boundaries of scale and sophistication, the role of DL compilers in bridging the gap between software abstractions and hardware capabilities becomes increasingly vital. 

\section{Empirical Validation of Compiler Behaviors}
\label{sec:experiments}

The preceding sections have examined the architectural design, major components, and optimization techniques of modern DL compilers. While these techniques aim to improve computational efficiency, their practical effectiveness can vary considerably across different workloads and execution conditions. To complement the architectural and technical analysis, we conduct an empirical evaluation of representative compiler behaviors by addressing three research questions: (RQ1) Which computation graph patterns benefit most from compiler optimization? (RQ2) How does compilation overhead affect the overall performance benefits of compiled execution? (RQ3) What are the trade-offs between static specialization and dynamic-shape compilation? We investigate these questions using controlled PyTorch/TorchInductor experiments, focusing on execution latency, memory consumption, compilation cost, numerical agreement, and dynamic-shape handling. The evaluation is intended to characterize representative optimization behaviors rather than compare the overall performance of different DL compiler frameworks.

\subsection{Methodology}

We compare PyTorch eager execution with \texttt{torch.compile} using the TorchInductor backend. Five synthetic inference workloads isolate common graph patterns without dataset or preprocessing effects: (1) a chain of elementwise nonlinearities, (2) a two-layer MLP with a residual connection, (3) a two-layer convolution--batch-normalization block, (4) layer normalization followed by gated elementwise operations and a residual connection, and (5) multi-head scaled dot-product attention with input and output projections. All workloads use FP32, inference mode, identical inputs, and PyTorch's default ``highest'' float32 matrix-multiplication precision; thus no precision change is counted as a compiler gain.

Experiments run on one NVIDIA RTX A6000 GPU using Python 3.10.15, PyTorch 2.3.1, CUDA 12.1, cuDNN 8.9.2, and Triton 2.3.1. For each execution mode, we perform 10 warm-up iterations followed by 50 measured iterations. GPU latency is measured using synchronized CUDA events and reported as the median; the interquartile ranges were narrow (at most 2.9\% of the median). First-call cost is synchronized wall-clock time from invocation through compilation and execution in a fresh TorchInductor cache. Incremental peak memory is measured relative to allocated memory immediately before an invocation. We check the output of every compiled static workload against eager execution. 

\subsection{Static-shape results}

\begin{table*}[t]
    \centering
    \caption{Eager and TorchInductor results on an RTX A6000. Latency is the median of 50 iterations after 10 warm-ups. Memory is incremental peak allocated GPU memory.}
    \label{tab:empirical-static}
    \resizebox{\textwidth}{!}{%
    \begin{tabular}{lrrrrrrr}
        \hline
        \textbf{Workload} & \textbf{Eager (ms)} & \textbf{Inductor (ms)} & \textbf{Speedup} & \textbf{First call (s)} & \textbf{Eager mem. (MB)} & \textbf{Inductor mem. (MB)} & \textbf{Max abs. error} \\
        \hline\hline
        Elementwise chain & 1.101 & 0.134 & $8.21\times$ & 1.781 & 192.0 & 32.0 & $4.77\times10^{-7}$ \\
        MLP residual & 2.154 & 2.142 & $1.01\times$ & 1.071 & 64.0 & 40.0 & $1.67\times10^{-6}$ \\
        Convolution block & 0.820 & 0.734 & $1.12\times$ & 2.464 & 75.0 & 75.6 & $5.48\times10^{-5}$ \\
        Norm residual & 0.626 & 0.141 & $4.43\times$ & 0.706 & 128.0 & 32.0 & $1.43\times10^{-6}$ \\
        Attention block & 3.140 & 3.150 & $1.00\times$ & 0.552 & 96.0 & 64.0 & $2.38\times10^{-7}$ \\
        \hline
    \end{tabular}}
\end{table*}

\begin{figure*}[t]
    \centering
    \includegraphics[width=0.88\textwidth]{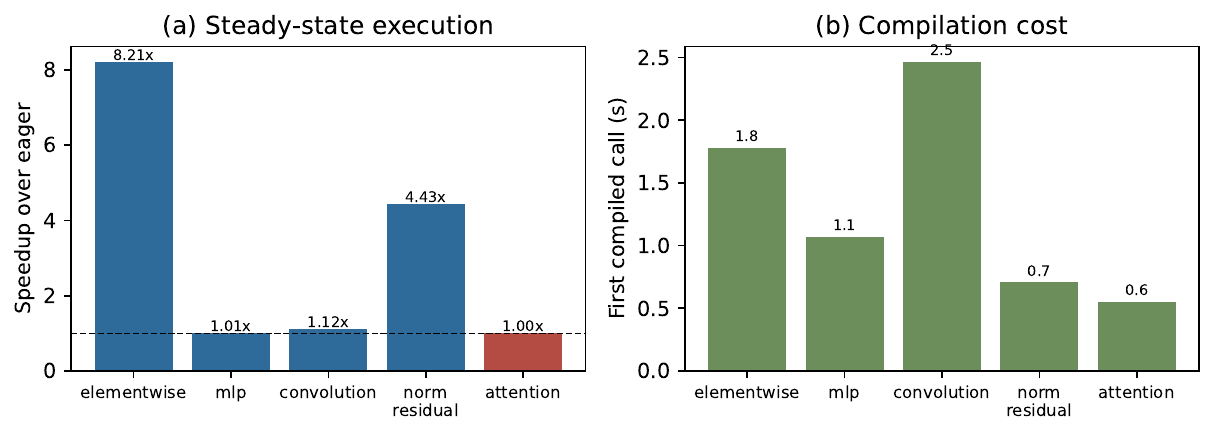}
    \caption{Steady-state speedup and synchronized first-call cost. Compilation is most effective when it can eliminate materialized intermediates; it does not accelerate every graph.}
    \label{fig:experimental-static}
\end{figure*}

Table~\ref{tab:empirical-static} and Figure~\ref{fig:experimental-static} show that graph structure determines the available optimization opportunity. The elementwise chain improves by $8.21\times$ and reduces incremental peak memory by 83.3\%, while the normalization-residual graph improves by $4.43\times$ and reduces memory by 75.0\%. These graphs expose several pointwise operations and intermediates that Inductor can fuse. The convolution block improves by a more modest $1.12\times$, and its memory footprint remains essentially unchanged. The MLP and attention blocks remain within 1\% of eager latency because their runtime is dominated by already optimized matrix-multiplication and scaled-dot-product-attention kernels; surrounding fusion does not materially shorten those kernels. This negative result is important: enabling a DL compiler is not itself evidence of a performance gain.

Compilation cost also changes the deployment decision. Dividing first-call cost by the steady-state time saved gives approximate break-even points of 1,843 calls for the elementwise chain, 1,458 calls for the normalization-residual graph, and 28,626 calls for the convolution block. The near-zero MLP improvement would require roughly 87,000 calls, while the slightly slower attention case has no latency break-even under this configuration. Ahead-of-time compilation, persistent caches, batching, and long-running services can amortize these costs; short-lived or latency-sensitive processes may not.

All compiled outputs agree closely with eager execution: the largest absolute deviation across the suite is $5.48\times10^{-5}$. Large pointwise relative errors can occur only where the reference is near zero, so we report maximum absolute error as the more interpretable conservative check. These tests establish numerical agreement for the sampled inputs but are not a formal correctness proof.

\subsection{Dynamic shapes}

To isolate shape handling, we run the elementwise graph at four previously unseen leading dimensions (2,304, 3,584, 5,632, and 7,168), holding the second dimension at 1,024. With static specialization, every new shape triggers a separate compile, totaling 2.529~s of first-call time. Symbolic dynamic compilation costs 0.757~s for the first shape and less than 0.2~ms for each subsequent first call, reducing cumulative shape-transition cost by approximately 70\%. The reusable dynamic kernel is 3.2--10.8\% slower in steady state than the corresponding specialized kernel.

\begin{figure}[H]
    \centering
    \includegraphics[width=0.78\columnwidth]{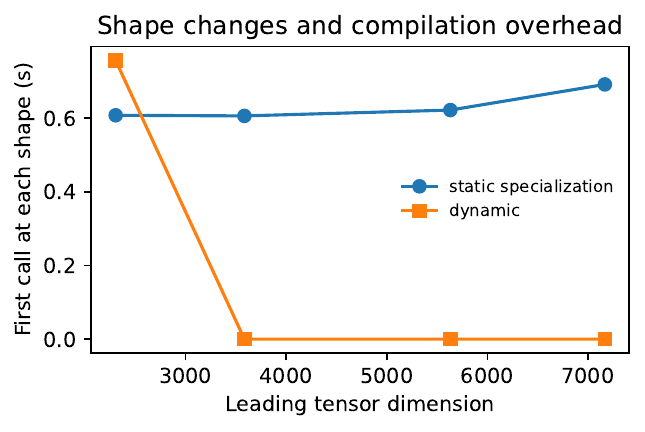}
    \caption{First-call time at four input shapes. Static specialization recompiles for each unseen shape, whereas one symbolic dynamic graph is reused.}
    \label{fig:experimental-dynamic}
\end{figure}

The dynamic-shape result exposes a recurring design tension: specialization can improve steady-state execution, while symbolic representations reduce recompilation and deployment disruption when shapes vary. The appropriate policy therefore depends on the shape distribution and expected reuse, rather than on a universally superior compilation mode.

\subsection{Scope and limitations}

The study deliberately favors transparency and reproducibility over breadth. It uses one GPU, one framework/compiler version, synthetic operator graphs, inference only, and a single precision. Results should not be interpreted as a leaderboard among TVM, XLA, TensorRT, or other systems, nor generalized to end-to-end models, training, other hardware, or other software versions. GPU clock variation and shared-server activity remain possible sources of noise, although narrow interquartile ranges reduce this concern. A broader evaluation would add multiple accelerators, end-to-end models, training workloads, energy, compile-cache behavior across processes, and multiple compiler stacks. Within these boundaries, the experiment provides direct evidence for the proposed relationships among fusion opportunity, amortization, memory behavior, and dynamic-shape specialization.

\section{Innovative future directions in DL compilers}

\noindent As deep learning applications grow in complexity and scale, the need for efficient, adaptable, and robust DL compilers becomes increasingly critical. Existing systems, while effective, face challenges such as framework-specific constraints, limited hardware support, and reliance on manual optimization, which hinder their ability to meet the diverse needs of modern neural network workloads. To address these limitations and unlock new capabilities, researchers are exploring innovative techniques that enhance efficiency, interoperability, and trustworthiness. This section systematically outlines several key research directions, systematically addressing these challenges with emerging techniques and their potential impact.

\textit{\textbf{Unified Framework and Hardware Integration}}.
Although there are compilers supporting more and more frameworks and hardware, there still exist problems, such as limited support for advanced operators, custom layers, and diverse architectures, which restrict portability and adoption across complex workloads. While ONNX has made strides in standardizing model formats, expanding support for advanced operators and custom layers across frameworks and hardware will enable broader usability. Future DL compilers should aim to seamlessly integrate with multiple frameworks and support diverse model representations on heterogeneous hardware. CMLCompiler focus on unifying the compilation process across diverse classical machine learning (CML) frameworks and algorithms  based on two unified abstractions, operator representations and extended computational graphs ~\cite{wen2023cmlcompiler}. A novel DL framework, NEST compiler, is proposed to efficiently deploy models across various AI accelerators by leveraging quantization, dynamic graph partitioning, and multi-level IR integration~\cite{park2024nest}.

\textit{\textbf{Machine Learning-Driven Optimization}}
Integrating machine learning (ML) techniques into DL compilers offers a powerful approach to automate optimization tasks, tackling the inefficiency of hand-crafted heuristics that struggle to scale across diverse workloads and hardware. Using methods such as reinforcement learning and neural architecture search (NAS), compilers can dynamically optimize operator fusion, scheduling, and kernel configurations with minimal human effort. CompilerDream, a model-based reinforcement learning approach, comprises a compiler world model and an agent to optimize code generation, highlighting the potential of differentiable programming techniques in compiler design~\cite{deng2024compilerdream}.

\textit{\textbf{Robustness and Security Enhancements}}
Enhancing debugging, verification, and privacy-preserving capabilities in DL compilers is critical to ensure reliability, correctness, and data security, particularly for sensitive applications where complexity, unverifiable outputs, and privacy risks erode trust. Debugging tools like HirGen expose errors in high-level IR optimizations—the most error-prone stage—offering insights into transformations and runtime behaviors~\cite{ma2023fuzzing}, and NNSmith uses fuzz testing to uncover bugs in frameworks like TVM, TensorRT, ONNXRuntime, and PyTorch, bolstering reliability~\cite{liu2023nnsmith}.
Verification methods, such as an SMT-based approach to validate multi-level IR translations~\cite{bang2022smt}, ensure correctness across compilation stages, complemented by formal verification to enforce adherence to specifications.
Privacy-preserving techniques further advance this effort, with HEIR introducing IRs for encrypted computations via homomorphic encryption~\cite{bian2023heir}. and SecureTVM automating secure neural inference through model translation~\cite{huang2023securetvm}, collectively fostering trust and transparency for mission-critical domains like healthcare and finance.

\textit{\textbf{Integration of Large Language Models}}
The integration of large language models (LLMs) into DL compiler workflows leverages their ability to understand and manipulate complex code structures, addressing the time-intensive nature of manual optimization and code generation that limits scalability. LLMs excel at suggesting optimizations, generating efficient machine-level instructions, and enhancing automation, as seen in Meta’s exploration of LLMs tailored for compiler tasks~\cite{cummins2024meta}. LLMs also optimize unstructured spaces and generate code templates for tensor accelerators~\cite{hong2024llm}, while WhiteFox, the first white-box compiler fuzzer, employs a multi-agent LLM framework with source-code context to detect deep logic bugs in DL compiler optimizations~\cite{yang2024whitefox}. This integration streamlines tedious tasks, boosts optimization creativity, and improves debugging, making compilers more efficient and accessible to a broader range of developers.

\section{Conclusion}

In this paper, we present a systematic analysis of deep learning compilers, covering their architectural evolution, compilation workflows, optimization techniques, and emerging research directions. By examining the frontend, middle-end, and backend stages, we provide a structured perspective on how modern DL compilers transform high-level model representations into efficient hardware-specific implementations. We further investigate representative optimization techniques and recent advances to highlight their roles in improving compiler efficiency and adaptability. To complement this analysis, we conduct an empirical evaluation of compiler behaviors across representative computation graphs. Our results demonstrate that compiler effectiveness strongly depends on workload characteristics, with fusion-friendly graphs achieving speedups of up to 8.21×, while workloads dominated by optimized library kernels exhibit limited performance improvements. We also show that compilation overhead may require thousands of executions to amortize and that dynamic-shape compilation can substantially reduce recompilation costs at the expense of modest steady-state performance degradation. These findings highlight the importance of considering workload structure, compilation cost, and shape variability when evaluating and deploying DL compilers. Looking forward, advances in cross-framework interoperability, machine learning-driven optimization, debugging and verification, privacy-preserving compilation, and LLM-assisted compiler workflows will continue to improve the capabilities of DL compilers. By combining systematic architectural analysis with empirical evidence, this work provides a reference for understanding existing compiler technologies and guiding future research and practical deployment.

\bibliographystyle{ACM-Reference-Format}
\bibliography{refer}
\end{document}